\documentclass[twoside,twocolumn]{article}

\usepackage{blindtext} % Package to generate dummy text throughout this template 

\usepackage[sc]{mathpazo} % Use the Palatino font
\usepackage[scale=0.9284]{tgpagella} % Adjusting font size
\usepackage[T1]{fontenc} % Use 8-bit encoding that has 256 glyphs
\usepackage{microtype} % Slightly tweak font spacing for aesthetics

\usepackage[english]{babel} % Language hyphenation and typographical rules

\usepackage[hmarginratio=1:1,top=32mm,columnsep=20pt]{geometry} % Document margins
\usepackage[hang, small,labelfont=bf,up,textfont=it,up]{caption} % Custom captions under/above floats in tables or figures
\usepackage{booktabs} % Horizontal rules in tables

\usepackage{lettrine} % The lettrine is the first enlarged letter at the beginning of the text

\usepackage{enumitem} % Customized lists
\setlist[itemize]{noitemsep} % Make itemize lists more compact

\usepackage{abstract} % Allows abstract customization
\usepackage{titlesec} % Allows customization of titles
\renewcommand\thesection{\Roman{section}} % Roman numerals for the sections
\renewcommand\thesubsection{\roman{subsection}} % roman numerals for subsections
\titleformat{\section}[block]{\large\scshape\centering}{\thesection.}{1em}{} % Change the look of the section titles
\titleformat{\subsection}[block]{\large}{\thesubsection.}{1em}{} % Change the look of the section titles

\usepackage{fancyhdr} % Headers and footers
\usepackage{titling} % Customizing the title section

\usepackage[dvipsnames]{xcolor}
\usepackage{hyperref}

\colorlet{mylinkcolor}{Black}
\colorlet{mycitecolor}{Black}
\colorlet{myurlcolor}{Blue}

\hypersetup{
  linkcolor  = mylinkcolor,
  citecolor  = black,
  urlcolor   = myurlcolor,
  colorlinks = true,
}

\usepackage{graphicx}
\pretitle{\begin{center}\huge\bfseries} % Article title formatting
\posttitle{\end{center}} % Article title closing formatting
\title{Brief Notes on Hard Takeoff, Value Alignment, and Coherent Extrapolated Volition} % Article title
\author{
\textsc{Gopal P. Sarma\href{http://orcid.org/0000-0002-9413-6202}{\includegraphics[scale=.10]{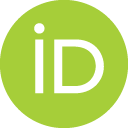}\textsuperscript{1}}\thanks{Email: gopal.sarma@emory.edu}\hspace{2pt}} \\ % Your name
\normalsize 1. \emph{Emory University, Atlanta, GA USA}\\ % Your institution
}
\date{} % Leave empty to omit a date
\renewcommand{\maketitlehookd}{%
\begin{abstract}
I make some basic observations about hard takeoff, value alignment, and coherent extrapolated volition, concepts which have been central in analyses of superintelligent AI systems.  
\end{abstract}
}

\begin{document}

% Print the title
\maketitle

%----------------------------------------------------------------------------------------
%	ARTICLE CONTENTS
%----------------------------------------------------------------------------------------

\section{On Hard Takeoff}
\setlength{\parindent}{0cm}
The distinction between \emph{hard takeoff} and \emph{soft takeoff} has been used to describe different possible scenarios following the arrival of human-level artificial intelligence.  The basic premise underlying these concepts is that software-based agents would have the ability to improve their own intelligence by analyzing and rewriting their source code, whereas biological organisms are significantly more restricted in their capacity for self-improvement \cite{bostrom2014superintelligence, good1965speculations, chalmers2010singularity, shanahan2015technological, shulman2010implications}.  \\

There is no precise boundary between the two scenarios, but in broad strokes, a hard takeoff refers to a transition from human level intelligence to superintelligence in a matter of minutes, hours or days.  A soft takeoff refers to a scenario where this transition is much more gradual, perhaps taking many months or years.  The practical importance of this qualitative distinction is that in a soft takeoff, there may opportunities for human intervention in the event that the initial AI systems have problematic design flaws. \\

The purpose of this brief note is simply to point out the following: takeoff speed refers to the rate of change of the agents' \emph{level of intelligence}, and not our perceived changes in the world around us.  Because the notion of an ``intelligence explosion''  has been constructed in analogy to a physical explosion \cite{yudkowsky2008artificial}, it gives rise to an inaccurate mental picture in people's minds. If self-improving AI systems\footnote{As this analysis neither requires nor implies that the driving force of change is a unitary agent, I have chosen to use the plural terms ``software-based agents,'' ``AI systems," or ``superintelligent AI systems.''  It may very well be a collection of agents / systems possessing powerful AI capabilities in aggregate.} are thought to be the intellectual analogue of nuclear chain reactions, then the natural image of an intelligence explosion that this metaphor creates is a scenario in which massive, disruptive changes take place in the world that are difficult for individuals and for society to handle.  \\

However, the premise of intelligent agents with capacities in substantial excess of any human being, which are able to process the sum total of human knowledge in the form of books, video, and ongoing contemporary events implies that greater levels of intelligence in the AI systems will be accompanied by actions taken with a corresponding level of information, insight, and operational skill.  Therefore, if the initial systems are designed correctly with respect to value alignment and goal structure stability, it is in fact a hard takeoff scenario which would be \emph{less disruptive} than a soft takeoff, not the other way around.  \\

I reiterate this claim for emphasis: \textbf{\emph{The takeoff speed of an intelligence explosion refers to the rate of change of intelligence in the AI systems, and not our perceived changes in the world around us.  Therefore, under the assumption of correctly designed systems, a hard takeoff is preferable to a soft takeoff because the resultant changes that take place in the world will be executed with greater precision, thoughtfulness, and insight.}}

\section{On Value Alignment and Coherent Extrapolated Volition} 
\setlength{\parindent}{0cm}
The preceding argument relied on a key assumption, namely that the AI systems capable of self-improvement were designed correctly with respect to value alignment and goal structure stability.  Value alignment refers to the construction of systems that take actions consistent with human values.  Russell states 3 design principles which encapsulate the notion of value alignment \cite{russell2016should}:
\begin{enumerate}
\item The machine's purpose must be to maximize the realization of human values. In particular, it has no purpose of its own and no innate desire to protect itself.
\item The machine must be initially uncertain about what those human values are.  The machine may learn more about human values as it goes along, but it may never achieve complete certainty.
\item The machine must be able to learn about human values by observing the choices that we humans make.
\end{enumerate}

A related notion is Yudkowsky's ``coherent extrapolated volition'' \cite{yudkowsky2004coherent, tarleton2010coherent}.  The basic premise of this proposal is that sophisticated AI systems will be capable of extrapolating and resolving disagreements between the value systems of individuals and groups, ultimately arriving at a goal structure that represents the collective desires of humanity.  This process of iterated reflection is analogous to Rawls' ``reflective equilibrium'' \cite{rawls}.  \\

Like the metaphor of hard takeoff, the notions of value alignment and coherent extrapolated volition can also give rise to an inaccurate mental picture, namely, that the aligned goal structure would either require or result in all humans arriving at complete agreement on all issues.  However, with adequate resources, it may very well be that value-aligned AI systems shape a world in which groups of individuals co-exist who disagree about object-level issues.  Certainly we can point to many examples in contemporary human society where individuals maintain divergent preferences without conflict. \\
\vspace{-4pt}

The purpose of this brief note is simply to point out the following:  \emph{\textbf{Implicit in any practical analysis of value alignment are the physical resources available to the AI systems.  In particular, the construction of a human compatible goal structure does not mean that all human disagreements have been resolved.  Rather, it means that a mutually satisfactory set of outcomes has been achieved, subject to resource constraints}}. \\
\vspace{-4pt}

It may be impossible to arrive at a consensus goal structure without adequate resources.  As a trivial example, if we have two individuals each of whom desires at least one apple, there is no disagreement if we have two apples.  On the other hand, if there is only one apple, conflict may very well be inevitable in the absence of other factors with which to resolve the imbalance between specific individual desires and total available resources.  In the context of superintelligent AI systems capable of exerting substantial influence on the world and shaping society on a global scale, adequate analysis of value alignment requires an understanding of the sum total of physical resources that the AI systems have at their disposal.  

\vspace{-8pt}
\subsection*{Acknowledgements}
I would like to thank Eric Drexler and Anders Sandberg for valuable discussions and feedback on the manuscript.  

\vspace{-8pt}
\subsection*{ORCID}

Gopal P. Sarma \raisebox{-.26\height}{\includegraphics[scale=.10]{orcid128}} \href{http://orcid.org/0000-0002-9413-6202}{0000-0002-9413-6202}\\

\bibliographystyle{ieeetr}
\bibliography{brief_notes_hard_takeoff}

\end{document}